\documentclass[11pt,a4paper]{article}
\usepackage{times,latexsym}
\usepackage{url}
\usepackage{graphicx}
\usepackage{booktabs}
\usepackage[T1]{fontenc}

\usepackage[acceptedWithA]{tacl2021v1}

\usepackage{amsmath,amsfonts,bm}

\def\eqref#1{equation~\ref{#1}}

\def\1{\bm{1}}

\DeclareMathAlphabet{\mathsfit}{\encodingdefault}{\sfdefault}{m}{sl}
\SetMathAlphabet{\mathsfit}{bold}{\encodingdefault}{\sfdefault}{bx}{n}

\usepackage{cleveref}

\usepackage{xspace,mfirstuc,tabulary}

\newif\iftaclinstructions
\taclinstructionsfalse 
\iftaclinstructions
\renewcommand{\confidential}{}
\renewcommand{\anonsubtext}{(No author info supplied here, for consistency with
TACL-submission anonymization requirements)}
\newcommand{\instr}
\fi

\iftaclpubformat 

\else

\fi

\title{i-Code V2: An Autoregressive Generation Framework \\over Vision, Language, and Speech Data}

\author{
Ziyi Yang, Mahmoud Khademi, Yichong Xu, Reid Pryzant, Yuwei Fang, Chenguang Zhu, \\ \bf Dongdong Chen, Yao Qian, Mei Gao, Yi-Ling Chen, Robert Gmyr, Naoyuki Kanda,\\
\bf Noel Codella, Bin Xiao, Yu Shi, Lu Yuan, Takuya Yoshioka, Michael Zeng, Xuedong Huang \\
Microsoft Azure Cognitive Services Research\\
\small \{\tt ziyiyang,chezhu\}@microsoft.com
}

\date{}

\begin{document}
\maketitle
\begin{abstract}
The convergence of text, visual, and audio data is a key step towards human-like artificial intelligence, however the current Vision-Language-Speech landscape is dominated by encoder-only models which lack generative abilities. We propose closing this gap with i-Code V2, the first model capable of generating natural language from any combination of Vision, Language, and Speech data. i-Code V2 is an integrative system that leverages state-of-the-art single-modality encoders, combining their outputs with a new modality-fusing encoder in order to flexibly project combinations of modalities into a shared representational space. Next, language tokens are generated from these representations via an autoregressive decoder. The whole framework is pretrained end-to-end on a large collection of dual- and single-modality datasets using a novel text completion objective that can be generalized across arbitrary combinations of modalities. i-Code V2 matches or outperforms state-of-the-art single- and dual-modality baselines on 7 multimodal tasks, demonstrating the power of generative multimodal pretraining across a diversity of tasks and signals.
\end{abstract}

\section{Introduction}
Pretrained Large language models (LLMs) have experienced massive success as general-purpose solutions for multiple tasks \cite{brown2020language}. However, a large gap persists between the capabilities of LLMs and true humanlike intelligence. This is partially because humans perceive a variety of sensory inputs while LLMs are typically restricted to Language (L) data and unable to understand or generalize to other modalities such as Vision (V) and Speech audio (S). 

Recently, the field of multimodal AI, which aims to develop AI systems capable of modeling multiple kinds of signals, has witnessed significant progress including new learning techniques \citep{radford2021learning,beit,alayracflamingo}, training data \citep{schuhmann2022laionb,zellers2022merlot,yang2022code}, and model architectures \citep{VL-BERT,visualbert,xu2022bridge}.

Despite this progress in multimodal AI, most research has focused on understanding pairs of modalities, such as speech-language and vision-language, and the fast-growing subfield of triple-modality AI (Language, Vision, Speech) remains limited to encoder-only models \citep{vatt,reserve,yang2022code}.

This paper proposes i-Code V2, the first encoder-decoder generative model for the triple-modality setting. i-Code V2 can flexibly generate text from arbitrary combinations of Language, Vision, and Speech data. This model addresses three ongoing challenges within multimodal research. 

\textbf{First}, at the time of writing, all existing triple-modality models are encoder-only, i.e. they can conduct discriminative tasks such as multimodal classification but not generative ones like visual question answering or automatic speech recognition. i-Code V2 enables the model to generate content from multimodal signals, unlocking more diverse applications and improved discriminative performance \cite{he2022z}. 

\textbf{Second}, most existing triple-modality research leverages triple-modality data (i.e. video with subtitles and audio track). However, the three modalities in video data can be noisily aligned \citep{howto100m} which degrades downstream pretraining. Furthermore, the available high quality video data is several orders of magnitudes smaller in size than single- or dual-modality datasets. For example, the largest publicly available image-caption dataset LAION \citep{schuhmann2022laionb} has 5 billion pairs (335 billion text tokens) while the largest video dataset MERLOT has 180M videos \citep{vatt,zellers2022merlot} and only 5 billion text tokens. i-Code V2 proposes a novel method for efficiently leveraging these larger and higher-quality dual- and single- modality datasets within a triple-modality pretraining framework. We accomplish this with a new, generalized sequence-to-sequence pretraining objective which converts multiple secondary objectives into simple text completion problems. 

\textbf{Third}, multimodal tasks are diverse in settings and data formats, e.g. Automatic Speech Recognition (ASR), vision QA, sentiment analysis, etc. Existing techniques apply separate inference strategies to each problem type, adding complexity and overhead for practitioners. i-Code V2 unifies all tasks under its text completion framework, rendering multimodal inference and cross-task transfer easier for practitioners.

i-Code V2 is built on top of state-of-the-art single-modality models: the vision and speech modalities are encoded with single-modality encoder respectively. Next, these encoded features and text token embeddings are inputted to a joint vision-language-speech encoder, which merges the different modalities into a shared representational space. Last, a language decoder, conditioned on the joint encoder via a cross-attention mechanism, is trained to generate language tokens autoregressively. 

We evaluate i-Code V2 thoroughly on 7 datasets, including multimodal summarization, multimodal dialogue generation, multimodal sentiment analysis, vision QA, vision captioning, and ASR. Notably, i-Code V2 outperforms previous SOTA models on MSMO (multimodal summarization), Image Chat (multimodal dialogue generation), UR-FUNNY (multimodal sentiment analysis). Additionally, i-Code V2 exhibits highly competitive performance compared to specialized dual-modality models on vision QA, vision captioning, and ASR, suggesting the power of integrative multimodal pretraining.

In summary, our key contributions are threefold:
\begin{enumerate}
\item We propose i-Code V2, the first vision-language-speech generative model that can generate natural language from one-, two- or three-modality inputs of image, video, language and speech.

\item We propose to pretrain i-Code V2 on a large-scale uni- and dual-modality datasets with a novel cross-modality text completion framework. The pretraining data resource is of higher quality and of larger scale, compared with video data previously used in three-modality learning. Using a sequence-to-sequence training objective, as opposed to the modality-specific objectives and task heads common in the multimodal AI space, can be flexibly applied to arbitrary combinations of modalities while simplifying the training and inference process.

\item i-Code V2 shows state-of-the-art or competitive performance across several multimodal tasks and domains, including multimodal natural language generation, vision QA, vision captioning, ASR, video sentiment analysis, multimodal summarization and dialogue generation.
\end{enumerate}

\section{Related Work}
\textbf{Multimodal Learning.} This field studies extracting and incorporating information from vision, language, and speech modalities. A recent advance is unifying models of different modalities to the transformer. For example, representing vision and language with one multimodal transformer model has shown great performance in image caption \citep{wang2022git, alayracflamingo}, vision classification \citep{coca}, vision question answering \citep{coca,li2022blip}, etc. In these models, extracted image features \citep{chen2020uniter} or projections of image patches \citep{wang2022simvlm, coca} are fused together with text token embeddings, then input to a multimodal encoder to obtain unified representations for vision and language. Lately, there has been significant progress in unifying representations for vision, language, and speech modalities. Three-modality encoder are pretrained on video data \citep{zellers2022merlot, yang2022code} or dual-modality data pairs \citep{yang2022code}. Multimodal representations can be integrated by a late-stage multimodal fusion network \citep{yang2022code}, or integrated early at the input stage \citep{zellers2022merlot}.

\noindent \textbf{Natural Language Generation (NLG).} Language generation has achieved significant process in many fronts in the era of deep learning, including text summarization \citep{liu2019text, yang2020ted, lewis2020bart, zhang2020pegasus,he2022z}, dialogue generation \citep{bao2020plato, zhao2020knowledge, chatgpt}, question answering \citep{rajpurkar2016squad, yang2018hotpotqa}, machine translation \citep{wu2016google}, etc. The transformer encoder-decoder model \citep{vaswani2017attention} is a powerful sequence-to-sequence generation architecture, due to its ability to efficiently model global dependencies in input text sequence. The Generative Pretrained Transformer (GPT) model families, which is transformer decoder only, have shown exceptional abilities in multi-task learning \citep{radford2019language}, few-shot in context learning \citep{brown2020language}, following human instructions \citep{ouyang2022training}, etc. The recently released large language model (LLM) ChatGPT, trained with Reinforcement Learning from Human Feedback (RLHF), can generate human-like text responses in fields like programming, tutoring, and professional writing.

\noindent \textbf{Generative Multimodal Model.} Generating one modality from another modality or a combination of input modalities has been an active field of multimodal learning. For example, image captioning \citep{johnson2016densecap,wang2022git}, automatic speech recognition \citep{yu2016automatic,radford2022whisper}, text-to-image generation \citep{ramesh2021zero,ramesh2022hierarchical,saharia2022photorealistic,rombach2022high}, etc. Inspired by the success of unified sequence-to-sequence multitask learning on NLP \citep{2020t5, lewis2020bart}, several recent works propose to unify vision-language tasks with one homogeneous model architectures and schemes. For example, \citet{wang2022ofa,lu2022unified,tang2022unifying} unite vision and vision-language tasks, such as image classification, object detection, semantic segmentation, visual QA, document understanding, image generation, etc. \citet{huang2023language} recently proposed the Kosmos-1 model to generate text based on vision and text input.

Different from previous works, i-Code V2 can not only encode and merge inputs from vision, language, and speech modalities, but also generate natural language. Such an integrative framework unifies various tasks across multimodal summarization, multimodal sentiment analysis, speech recognition, visual QA and vision captioning.

\section{An Integrative Multimodal Generative Model}
In this section, first we will introduce the design and model architecture of i-Code V2. Then we present how we pretrain i-Code V2 on diverse large-scale multimodal datasets.

\subsection{Model Architecture}
i-Code V2 model consists of multimodal encoders and a language decoder. Following the spirit of integrative AI \citep{yang2022code}, the language, vision and speech modalities are encoded by their corresponding encoder or converted to numerical representations respectively, before fused with each other. Leveraging pretrained models enables us to utilize state-of-the-art model architecture for each modality. It is also computationally efficient since these models have already been extensively trained on single-modality data. This also gives us the flexibility of choosing preferred encoders. For example, we can use a medical-domain specific language encoder-decoder; or choose smaller speech/vision encoder on devices with low computation resource, without having to re-design the framework.

In detail, we leverage the following state-of-the-art single-modality encoders:

\paragraph{Vision Encoder.} In different multimodal scenarios, the vision modality can be either a single image or a video (especially when the input contains speech). To flexibly encode and represent the vision modality input, we opt to use OmniVL, a foundation model for both image-language and video-language \citep{wang2022omnivl}. It uses independent 2D/3D convolution-based patch tokenizer to first process image/video and a unified vision transformer to generate vision representations. It has 122 million parameters.

\paragraph{Speech Encoder.} We use WavLM large \citep{chen2022wavlm}, a speech encoder pretrained on 94k-hour data in a self-supervised manner. Pretraining objectives include  masked speech denoising and predicting. The model architecture is a transformer encoder with Gated Relative Position Bias on top of a temporal CNN-based featurizer. The parameter size is 315 million.

\paragraph{Joint Vision-Language-Speech Encoder.} We use a 24-layer transformer encoder to jointly encode vision, language and speech modalities. After the vision and speech modality inputs are encoded by their respective encoder, a 1-layer projection (one for each modality) transforms the features into the same dimension as the text vocabulary embedding. Transformed features are concatenated with the text tokens embeddings and then input into the transformer layers for both inter- and intra-modality attention.

We initialize the transformer layers of the joint encoder using the encoder part of the recently developed Z-Code++ summarization model, which has 485 million parameters and was pretrained using generative training objectives on 160G of English text data \citep{he2022z}.

\paragraph{Language Decoder with Multimodal Cross-Attention.}
i-Code V2 then uses a decoder to generate textual sequences from the multimodal encoder output. The 24-layer decoder cross-attends with the multimodal representation from the joint Vision-Language-Speech encoder. We use the pretrained transformer \emph{decoder} from the Z-Code++ model (485 million parameters) to initialize these parameters.

\begin{figure}
\centering
\includegraphics[width=0.8\columnwidth]{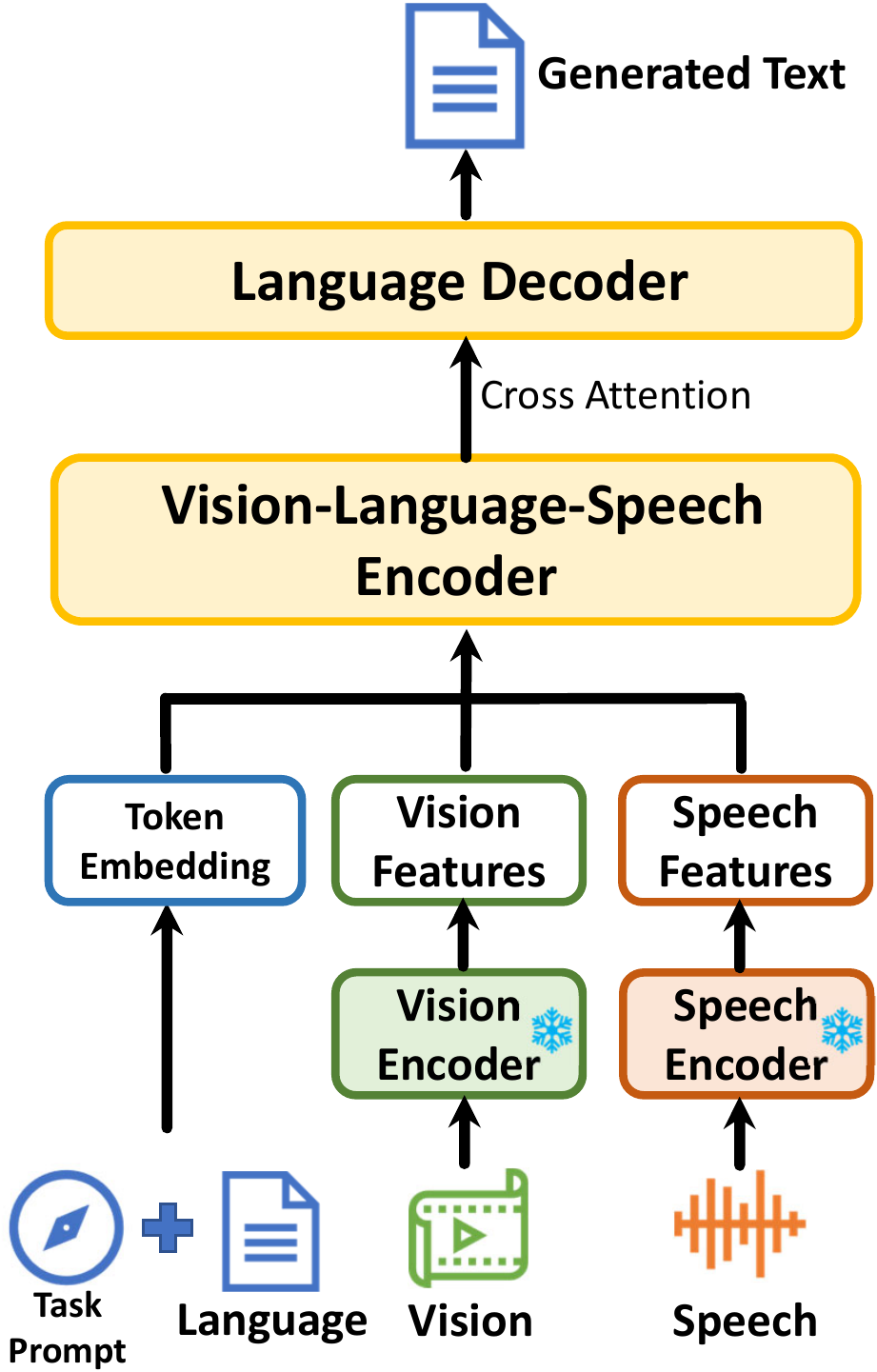}
\caption{i-Code V2 Model Architecture. Parameters of vision and speech encoders are frozen during pretraining and are updated in finetuning.}
\label{fig:model}
\end{figure}

\subsection{Large-Scale Multimodal Generative Pretraining}
\label{sec:pretrain}
We leverage a collection of large-scale dual modality datasets to conduct speech-language generative pretraining, vision-language generative pretraining, and language-language generative pretraining. In particular, our pretraining objectives adopt a simple sequence-to-sequence strategy, which poses each modality-specific and cross-modality objective as a text completion.

\subsubsection{Vision-Language Generative Pretraining}
\paragraph{Image Captioning.} Given an image, the model predicts the corresponding textual caption. We use the 72.8 million subset of Florence image-text pair dataset \citep{florence}. The task prompt is ``Generate the caption for this image: ''. 

\paragraph{Video Captioning.} The pretraining task is to generate the caption of a video clip. We use the largest-scale publicly available video captioning dataset WebVid-10M \citep{bain2021frozen}, which contains 10.7M video-caption pairs. The task prompt is ``Generate the caption for this video: ''.

\paragraph{Vision Question \& Answering.} For this task, we use VQA v2 training set, an open-ended vision question answering dataset \citep{antol2015vqa}, which has 443,757 question-answer pairs. The task prompt is ``Answer the following question based on the image: ''.

\paragraph{Vision-Augmented Text Reconstruction.} This pretraining task aims to improve the model's ability on cross-modal understanding. We mask spans of the textual image caption and replace them with sentinel tokens, similar to T5 pretraining \citep{2020t5}. The model needs to predict masked out text spans, given the masked textual input and the image. The data resource is the same as in ``Image Captioning''. The task prompt is ``Reconstruct the following text based on the image: ''.

\subsubsection{Speech-Language Generative Pretraining}
We leverage three sources of labeled data for generative speech-language learning:
\paragraph{Speech transcription.} This dataset contains 75k-hour human-transcribed speech utterances \citep{yang2022code}, collected from scenarios such as call center and AI voice assistant. The input is the speech utterance, and the target output is the transcription. The pretraining loss is the cross entropy between the target and prediction. The task prompt is ``Transcribe the speech utterance to text: ''.

\paragraph{Speech Sentiment Analysis.} The goal of this task is to predict the sentiment of a speech utterance, e.g., from ``highly negative'' to ``highly positive''. We gather data from CMU Multimodal Opinion Sentiment and Emotion Intensity (CMU-MOSEI) \citep{mosei} and Spoken Language Understanding Evaluation (SLUE) \citep{slue}. The task prompt is ``Predict the sentiment of this segment: ''. The output target is the textual sequence of the ``sentiment''.

\paragraph{Speech Emotion Recognition.} The task is to predict the emotion category of a speech utterance, including \{happiness, sadness, anger, fear, disgust, surprise\}. The dataset is from the emotion intensity subtask of CMU-MOSEI. The target generation sequence is the emotion category name. The task prompt is ``Predict the emotion of this segment: ''.

\paragraph{Speech-Augmented Text Reconstruction.} Similar to ``Vision-Augmented Text Reconstruction'', we mask spans of the speech transcription and ask the model to predict masked-out text spans, given the speech input as well. The task prompt is ``Reconstruct the following text based on the speech: ''.

\subsubsection{Language-only Generative Pretraining.} We include high-quality text-only corpus, i.e. English Wikipedia and BookCorpus \citep{zhu2015aligning}, in pretraining as a supplement to the language-modality data of vision-language and speech-language datasets. This language-only pretraining task follows T5 where the input is span-masked text, and the output is the original masked span. The task prompt is ``Reconstruct masked spans in the following text: ''.

\begin{figure*}
\centering
\includegraphics[width=\textwidth]{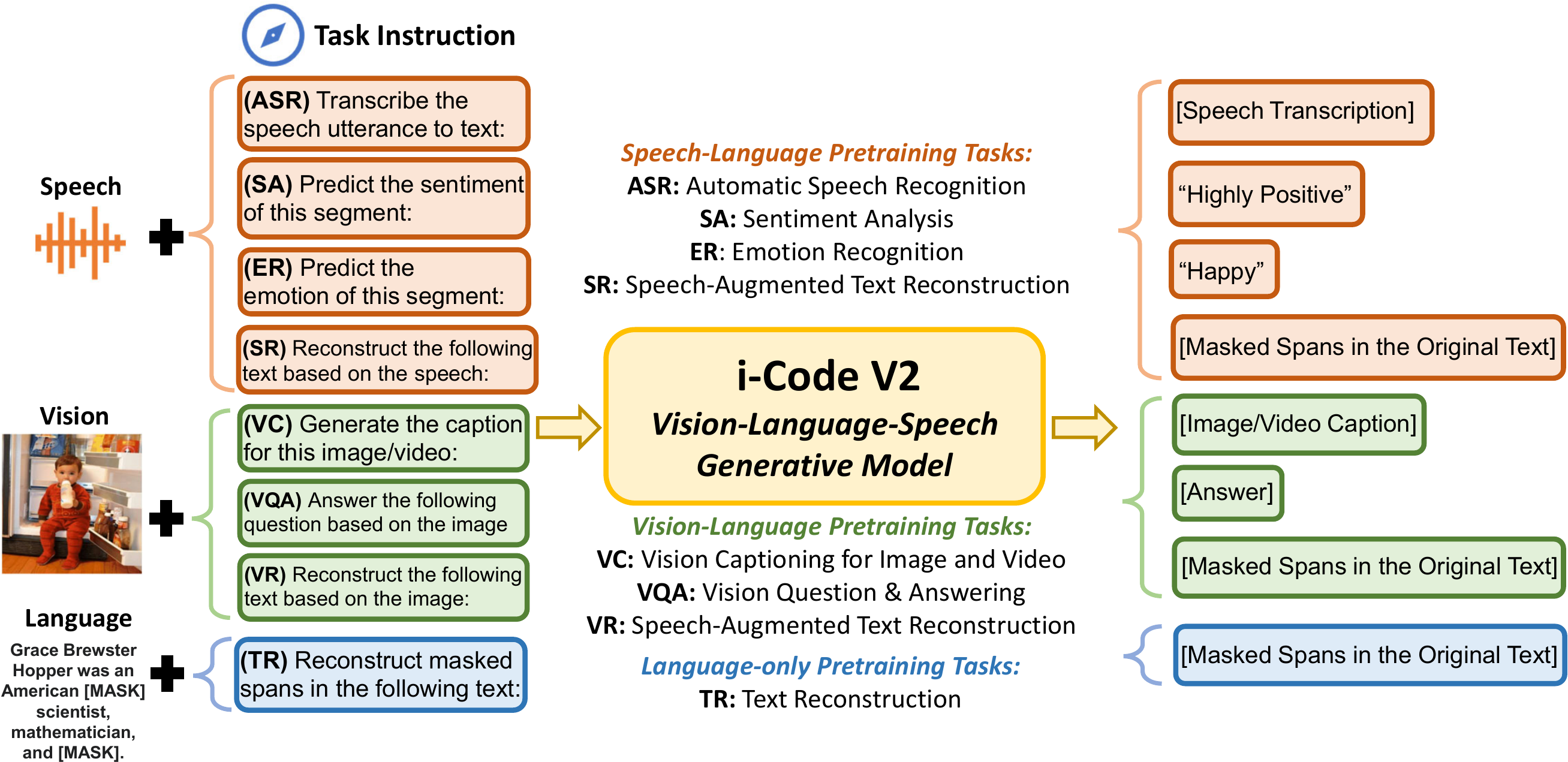}
\caption{i-Code V2 multimodal pretraining. It unifies tasks across vision, language and speech domains to text completion/generation training objectives. Pretraining tasks include both unimodal, e.g. TR, and dual-modal ones, e.g., ASR, SA, ER, SR, VC, VQA, and VR (full names of task initials are provided in the figure).}
\label{fig:multitask}
\end{figure*}

The multimodal pretraining process, task, and textual instructions are illustrated in \Cref{fig:multitask}.

\subsubsection{Pretraining Details}
 To expedite the pretraining process, we freeze the speech and vision encoders during pretraining, only updating the parameters of the Vision-Language-Speech encoder and the language decoder. For each optimization step, we select the pretraining dataset from the candidate pool using ``Exponentially Smoothed Weighting (ESW)'' sampling. ESW is widely used in multilingual pretraining \citep{devlin2019bert} where multilingual corpus sizes can be different with several magnitudes. Assume the size ratio of the dataset $A$ in the overall training datasets is $P(A)$. We exponentiate the ratio by the factor $S<1$ then we sample datasets according the re-normalized exponential ratio $\frac{p(A)^S}{\sum_{A}{p(A)^s}}$. We use $S=0.5$ in our setting.

 We pretrain the i-Code V2 model on datasets introduced above for 1 epoch on 24 A100 GPUs, with batch size 8 (per-GPU) and gradient accumulation steps 3. Having accumulation steps > 1 also makes the effective optimization batch contain data from different resources.

\section{Experiments}
We test i-Code v2 on 7 datasets from assorted categories, including multimodal summarization, multimodal dialogue generation, video multimodal sentiment analysis, ASR, vision QA, and vision captioning. In downstream tasks, we update parameters of Vision-Language-Speech encoder, language decoder, and single-modality encoders. Finetuning hyperparameters can be found in \Cref{sec:analysis}.

Overall, i-Code V2 sets a new state-of-the-art in 3 tasks (MSMO, Image Chat, and UR-FUNNY) and remains highly competitive in the rest, suggesting the promise of integrative and generative multimodal pretraining.

\subsection{Multimodal Summarization}
We first evaluate on multimodal summarization task. Well studied in the field of natural language processing, in traditional summarization, input only contains language. However, in many real-world scenarios, such as multimedia coverage and online news article, key information is also included other modalities, e.g., pictures. We test i-Code V2 on the multimodal news summarization dataset MSMO \citep{zhu2018msmo}. Given a news article with image(s), the task is to generate a few-sentence summarization. Its training/validation/test split contains 293,965/10,355/10,261 news articles with images from Daily Mail website. The ground-truth ``golden'' summary is the highlight written by the news editor. The evaluation metrics are ROUGE scores \citep{lin2004rouge}. Baseline models include:

\paragraph{Text-only summarization model.} BertSum \citep{liu2019text} model variants BertAbs and BertExtAbs, BART \citep{lewis2020bart}, ZCode++ \citep{he2022z}.

\paragraph{UniMS \citep{zhang2022unims}} An encoder-decoder multimodal summarization model that can process multimodal inputs and select images.

\paragraph{MOF \citep{zhu2020multimodal}} A multimodal generation model with the guidance of multimodal reference.

\paragraph{ATG/ATL/HAN} These are baselines from the original MSMO dataset paper, that Point Generator Network \citep{pgn} attending with global vision features(ATG), attending with local vision features (ATL), and hierarchical attention with local features.

The task prompt used in i-Code V2 is ``Summarize this article with the images: ''. As shown in \Cref{tab:msmo}, compared with baseline models, i-Code V2 has shown competitive performance on ROUGE-1 and ROUGE-L. Compared with the language encoder-decoder ZCode++, that i-Code V2 encoder-decoder is initialize from, i-Code V2 shows considerable improvement, which demonstrates the effectiveness of the proposed multimodal pretraining.

\begin{table}
\centering
\begin{tabular}{l|ccc}
\toprule
Model & R1 & R2 & RL \\ \midrule
BertAbs & 39.02 & 18.17 & 33.20  \\
BertExtAbs & 39.88 & 18.77 & 38.36 \\
BART & 41.83 & 19.83 & 39.74  \\
ZCode++ & 42.19 & 20.03 & 37.2 \\
UniMS & 42.94 & 20.50 & \textbf{40.96} \\
MOF (enc) & 41.05 & 18.29 & 37.74 \\
MOF (dec) & 41.20 & 18.33 & 37.80 \\
HAN & 40.82 & 18.30 & 37.70 \\
ATL & 40.86 & 18.27 & 37.75 \\
ATG & 40.63 & 18.12 & 37.53 \\
\textbf{i-Code V2} & \textbf{44.7} & \textbf{21.0} & 37.7 \\
\bottomrule
\end{tabular}
\caption{Results on the multimodal news summarization MSMO test set.\label{tab:msmo}}
\end{table}

\subsection{Multimodal Dialogue Generation}
Our model also has the ability perceive contextual multimodal signals to generate textual response. A type of datasets to measure such ability is multimodal-based conversation. For example, for the multimodal open-domain dialogue dataset Image-Chat \citep{shuster2020image}, each data example includes an image; the dialogue history between two speakers A and B; speaker style traits. The task objective is to generate the next-round dialogue. Baselines include:

\paragraph{BlenderBot \citep{roller2020recipes}} A sequence-to-sequence ChatBot model of 2.7 Billion parameters pretrained on 1.5B Reddit comment conversations. 

\paragraph{Multi-Modal BlenderBot\citep{shuster2021multi}} The multimodal version of BlenderBot that fuses vision features from ResNet/Faster-RCNN in the multimodal text generation.

\paragraph{2AMMC \citep{ju2019all}} A multimodal generative model that combines ResNet and text transformer.

\paragraph{DialoGPT \citep{zhang2019dialogpt}} A GPT model trained on 147 million social media dialogues.

 With the task prompt, we can conveniently guide the model to generate dialogue in the speaker style, the prompt is ``Generate the response for the dialogue in \{style type\} style: ''. For fair comparison, we do not include baseline model that co-trains on multiple multimodal dialogue datasets e.g., (\citet{shuster2020image}). The evaluation metric includes F1 and ROUGE-L. Results in \Cref{tab:image-chat} show that i-Code V2 has significantly outperformed previous baselines on both F1 and R-L metrics.

\begin{table}
\centering
\begin{tabular}{l|cc}
\toprule
Model & F1 & RL \\ \midrule
DialoGPT & 6.2 & 5.2 \\
2AMMC & 9.3 & 11.0 \\
BlenderBot & 9.2 & 12.3 \\
Multi-Modal BlenderBot & 13.1 & 18.0 \\
\textbf{i-Code V2} & \textbf{15.5} & \textbf{18.6} \\
\bottomrule
\end{tabular}
\caption{Results on the multimodal dialogue generation dataset Image Chat.\label{tab:image-chat}}
\end{table}

\subsection{Video Multimodal Sentiment Analysis}
We further evaluate on multimodal sentiment analysis datasets. E.g., in UR-FUNNY \citep{urfunny}, a humor detection dataset, the input is a video, the audio of the video, and the text transcript. The task is to predict whether the immediate laughter will follow the clip. The dataset contains 5306/1313/1638 humor instances for train/validation/test split, and 5292/1313/1652 for the not humor instances. Although previous models approached this problem as binary classification, we finetune i-Code V2 to directly predict the target sequence ``funny''/``unfunny'', with task prompt ``Predict the sentiment of this clip: ''. We compare i-Code V2 with baselines that use all three-modality inputs. i-Code V2 outperforms previous models by large margins. This shows that the multimodal encoder in i-Code V2 can effectively fuse signals of vision, language and speech modalities, and the decoder can successfully attend with the multimodal encoder outputs.

\begin{table}
\centering
\begin{tabular}{l|c}
\toprule
Model & Accuracy \\ \midrule
ZCode++ & 75.4 \\
MulT \citep{mult} & 70.55 \\
MISA \citep{misa} & 70.61 \\
MultiBench \citep{multibench} & 66.7 \\
BBFN \citep{bbfn} & 71.68 \\
LMF \citep{lmf} & 67.53 \\
TFN \citep{tfn} & 68.57 \\
\textbf{i-Code V2} & \textbf{79.59} \\
\bottomrule
\end{tabular}
\caption{Prediction accuracy on UR-FUNNY dataset.\label{tab:urfunny}}
\end{table}

\subsection{Automatic Speech Recognition}
Automatic Speech Recognition (ASR) transfers human-spoken language into text. As a multimodal generative model, i-Code is able to process speech signals and generate the text transcript in a sequence-to-sequence manner. We evaluate on the classical ASR dataset LibriSpeech \citep{panayotov2015librispeech}. We finetune i-Code V2 on LibriSpeech 960h training data and test on the test-clean split. We compare i-Code V2 with the following models:

\paragraph{WavLM \citep{chen2022wavlm}} Transformer-based speech encoder that is pretrained on audio data with self-supervised learning
\paragraph{wav2vec 2.0 \citep{baevski2020wav2vec}} A speech representation model with CNN-Transformer architecture, pretrained with a contrastive self-supervised task on quantized speech representations.
\paragraph{S2T Transformer \citep{wang2021fairseq}} A transformer-based text-to-speech model provided in the Fairseq \citep{ott2019fairseq} sequence modeling toolkit.  
\paragraph{Whisper \citep{radford2022whisper}} A recently developed speech recognition system that is pretrained on 680K hours labeled speech-text transcript with multitask-supervision.

\begin{table}
\centering
\begin{tabular}{l|c}
\toprule
Model & WER(\%)$\downarrow$ \\ \midrule
wav2vec 2.0 & 2.0 \\
WavLM Large & 2.1 \\
Whisper Large & 2.7 \\
Whisper Medium & 4.12 \\
S2T Transformer Large & 3.2 \\
\textbf{i-Code V2} & 3.86 \\
\bottomrule
\end{tabular}
\caption{Word Error Rate (WER) on LibriSpeech dataset test-clean split.\label{tab:librispeech}}
\end{table}

The task prompt is ``transcribe the speech utterance to text: ''. Results are presented in \Cref{tab:librispeech}. i-Code V2 exhibits competitive performance on this task compared with ASR models specifically designed for this task, showing it is capable of decoding speech signals to language. Note that WavLM Large result presented in \Cref{tab:librispeech} is using Connectionist temporal classification (CTC) decoding on top of the speech encoder, which is superficially designed for ASR task. While the language decoding in i-Code V2 is for general purpose.

\subsection{Vision QA}
An important domain in vision-language learning is vision question \& answering (QA). The task is given an image and a question about the image, the model needs to predict the answer. The evaluation benchmark we test on is the Visual Question Answering (VQA) 2.0 \citep{antol2015vqa}. The train/validation/test data split has 443,757/214,354/447,793 questions respectively. Previous vision-language works, including those with language-generation functionality, almost all convert this task into a classification task: the models are trained to the answer from 3129 most frequent candidates (e.g. \citep{wang2022image}). Unlike this closed-vocabulary setup, we adopt an open-vocabulary setting that i-Code V2 is trained to generate the answer. The task prompt is ``Answer the following question based on the image:''. Note that we don't provide candidate answer choices to i-Code V2 during testing. \Cref{tab:vqa} contains baselines for both settings. i-Code V2's performance is competitive compared with vision-language models such as VisualBERT, LXMERT and VL-BERT. It is worth noting that Flamingo is pretrained on 2.1B vision-language data examples and has 80B parameters. In comparison i-Code V2 is pretrained on < 80M vision-language data and only has 1.4\% parameters of Flamingo (1.1B parameters, vision encoder, multimodal encoder, and the language decoder).

\begin{table}
\centering
\begin{tabular}{l|c}
\toprule
Model & Accuracy \\ \midrule
\multicolumn{2}{c}{Closed-Vocabulary} \\ \midrule
VisualBERT \citep{li2020does} & 71.0 \\
LXMERT \citep{tan2019lxmert} & 72.5 \\
FLAVA & 72.8 \\
OSCAR & 73.16 \\
VL-BERT \citep{VL-BERT} & 72.2 \\
BLIP \citep{li2022blip} & 78.32 \\
CoCa \citep{coca} & \textbf{82.3} \\ \midrule
\multicolumn{2}{c}{Open-Vocabulary} \\ \midrule
Flamingo*\citep{alayracflamingo} & \textcolor{gray}{82.1} \\
\textbf{i-Code V2} & 75.10 \\
\bottomrule
\end{tabular}
\caption{Comparison with baselines on VQA 2.0 test set.\label{tab:vqa}}
\end{table}

We further test i-Code V2 on the VizWiz-VQA dataset \citep{gurari2019vizwiz}, which is designed to answer visual questions from visually impaired people. It has 20,000/3,173/8000 image-question-answer triplet for the train/validation/test data split. Baselines include VisWiz Challenge Winner \citep{liu2021enhancing}, BAN \citep{kim2018bilinear}, B-Ultra \& B-FRCNN \citep{changpinyo2019decoupled}. Our performance is obtained from submitting to VizWiz challenge server. Again, we adopt the open-vocabulary evaluation setting. i-Code V2 shows better performance than the previous VizWiz challenge winner and provides a strong baseline for models of intermediate size. As will be discussed in \Cref{sec:analysis}, i-Code V2 also shows impressive zero-shot performance on VizWiz-VQA dataset, given that it is an open-vocabulary generative model.

\begin{table}
\centering
\begin{tabular}{l|c}
\toprule
Model & Accuracy \\ \midrule
BAN & 51.6 \\
B-FRCNN & 51.9 \\
B-Ultra & 53.7 \\
LXMERT & 55.4 \\
VisWiz Challenge Winner & 60.6 \\
Flamingo* & \textcolor{gray}{65.4} \\
\textbf{i-Code V2} & \textbf{61.3} \\
\bottomrule
\end{tabular}
\caption{Performance on VisWiz-VQA test-std set.\label{tab:wiz}}
\end{table}

\subsection{Image Captioning}
Image captioning is an important field of multimodal understanding and generation. We evaluate i-Code V2 on the MS-COCO dataset \citep{chen2015microsoft} with the Karpathy test split \citep{karpathy2015deep}, with results presented in \Cref{tab:mscoco}. Evaluations metrics include BLEU@4, METEOR, CIDEr, and SPICE. The task prompt is ``Generate the caption for this image: ''. Baseline methods include image captioning models BUTD \citep{anderson2018bottom} and AoANet \citep{huang2019attention}; vision-language generative models, e.g., VL-BART, VL-T5 \citep{cho2021unifying}, XGPT\citep{xia2021xgpt}; models using additional auxiliary input such as UNITAB \citep{yang2022unitab} with object detection information. i-Code v2 outperforms vision-language baselines on METEOR, CIDEr, and SPICE.

\begin{table}
\centering
\resizebox{\columnwidth}{!}{
\begin{tabular}{l|cccc}
\toprule
Model & BLEU@4 & METEOR & CIDEr & SPICE \\ \midrule
VL-T5 & 34.5 & 28.7 & 116.5 & 21.9 \\
VL-BART & - & - & 116.6 & - \\
BUTD & 36.2 & 27.0 & 113.5 & 20.3 \\
AoANet & \textbf{37.2} & 28.4 & 119.8 & 21.3 \\
UNITAB & 35.8 & 28.4 & 119.1 & 21.5 \\
XGPT & 37.2 &  28.6 & 120.1 & 21.8 \\
\textbf{i-Code V2} & 36.8 & \textbf{28.9} & \textbf{124.3} & \textbf{22.3} \\
\bottomrule
\end{tabular}
}
\caption{Experimental results (with cross-entropy optimization) on MS-COCO image captioning dataset (Karpathy test split).\label{tab:mscoco}}
\end{table}

\subsection{Analysis \& Explorations}
\label{sec:analysis}
\paragraph{Ablation study on pretraining effectiveness.} We investigate the pretraining effectiveness by comparing performance of i-Code V2 with and without the multimodal pretraining (\Cref{sec:pretrain}). As shown in \Cref{tab:pretrain_ablation}, the multimodal pretraining further improves the performance on downstream tasks. The improvement is more significant on tasks where cross-modality understanding is more crucial, such as video sentiment analysis.

\begin{table}
\centering
\resizebox{\columnwidth}{!}{
\begin{tabular}{l|c|cc|c}
\toprule
variant & UR-FUNNY & \multicolumn{2}{c|}{Image Chat} & LibriSpeech \\ \midrule
i-Code V2 & Accuracy & F1 & R-L & WER(\%)$\downarrow$\\ \midrule
 w/ pretraining & \textbf{79.59} & \textbf{15.5} & \textbf{18.6} & \textbf{3.86} \\
 w/o pretraining & 62.85 & 15.0 & 18.2 & 12.1 \\
\bottomrule
\end{tabular}
}
\caption{Ablation study of the proposed multimodal pretraining.\label{tab:pretrain_ablation}}
\end{table}

\paragraph{Ablation study on pretraining objectives.} We explore how pretraining objectives affect the model performance. For example, as shown in \Cref{tab:objective_ablation}, removing ``Language-only Generative Pretraining'' (LGP) is adversarial for the performance on multimodal summarization dataset MSMO, while it has negligible effect on ASR.

\begin{table}
\centering
\resizebox{\columnwidth}{!}{
\begin{tabular}{l|ccc|c}
\toprule 
Dataset & \multicolumn{3}{c|}{MSMO} & ASR \\ \midrule
Metric & R1 & R2 & RL & WER ($\downarrow$) \\ \midrule 
w/o LGP & 42.23 & 20.12 & 37.1 & 3.88 \\
Full Pretraining & 44.7 & 21.0 & 37.7 & 3.86 \\ 
\bottomrule
\end{tabular}
}
\caption{Ablation study on pretraining objectives. LGP stands for ``Language-only Generative Pretraining''.\label{tab:objective_ablation}}
\end{table}

\paragraph{Zero-shot Learning.} We further explore i-Code V2 transfer learning ability. We test pretrained i-Code V2 directly on VizWiz-VQA without finetuning. As a open-vocabulary generative model, the zero-shot performance of i-Code V2 is respectful, with overall accuracy 22.53\%, and 73.6\% for ``Yes/No'' answers, 8.47\% for ``Number'' answers, 24.46\% for ``Other'' answers, 10.49\% for ``Unanswerable'' answers respectively (for reference, Flamingo-9B zero-shot accuracy is 28.8\%). Note that VizWiz-VQA questions are from visually impaired population and images are also distinct from those in VQA 2.0 that is in the pretraining data. With the significant difference in dataset distribution between VizWiz-VQA and VQA 2.0, this performance indicates that i-Code V2 can closely follow the task instruction to answer the question (e.g., other than generating the caption). It also shows that i-Code V2 learns to answer visually grounded questions from pretraining, even though there are assorted pretraining tasks and datasets.

\paragraph{Training Hyperparameters for Downstream Tasks.} In \Cref{tab:ft_hyper}, we list the learning rate, batch size (per GPU), and epochs for each finetuning dataset. We choose the finetuning checkpoint with the best performance on the validation test for the final evaluation. All finetuning jobs are conducted on eight A100 GPUs.

\begin{table}
\centering
\resizebox{\columnwidth}{!}{
\begin{tabular}{l|ccc}
\toprule
Task & lr & Batch Size & Epochs \\ \midrule
MSMO & $2.5\times 10^{-5}$ & 8 & 12 \\
Image Chat & $2\times 10^{-5}$ & 8 & 5 \\
UR-FUNNY & $1\times 10^{-5}$ & 8 & 12 \\
LibriSpeech & $2\times 10^{-5}$ & 2 & 10 \\
VQA &$2\times 10^{-6}$  &  2& 4 \\
VisWiz-VQA &$2\times 10^{-6}$ & 2 & 4 \\
MS-COCO &$2\times 10^{-6}$&  2&  4\\
\bottomrule
\end{tabular}
}
\caption{Training hyperparameters on downstream tasks.\label{tab:ft_hyper} }
\end{table}

\section{Conclusion}
In this paper, we propose i-Code V2, a multimodal generative model that jointly encodes language, vision and speech modalities and decodes the corresponding natural language sequence. i-Code V2 is pretrained on assorted high-quality single- and dual-modality datasets, where different tasks are unified as a multimodal sequence-to-sequence generation paradigm. i-Code V2 exhibits impressive performance in various multimodal generation domains, including multimodal nature language generation, ASR, vision QA, vision captioning and video sentiment analysis.

\appendix

\bibliography{tacl2021}
\bibliographystyle{acl_natbib}

\end{document}